\documentclass[pmlr,twocolumn,10pt]{jmlr} 

\jmlrvolume{259}
\jmlryear{2024}
\jmlrsubmitted{LEAVE UNSET}
\jmlrpublished{LEAVE UNSET}
\jmlrworkshop{Machine Learning for Health (ML4H) 2024} 

\usepackage[switch]{lineno}

\usepackage{booktabs}       
\usepackage{multirow}
\usepackage{amsfonts}       
\usepackage{nicefrac}       
\usepackage{microtype}      
\usepackage{xcolor}         
\usepackage{comment}        
\usepackage{verbatim}       
\usepackage{algorithm}
\usepackage[noend]{algorithmic}
\usepackage{makecell}
\usepackage{siunitx}
\usepackage{mwe}

\usepackage{amssymb}
\usepackage{amsmath}
\usepackage{bm, bbm}
\usepackage{mathtools}
\usepackage{enumitem}
\setlist{itemsep=0.2ex, topsep=0ex, partopsep=0ex, parsep=0.5ex, leftmargin=2.3ex}
\usepackage[skip=1pt]{caption}

\def\R{\mathbb{R}}
\def\Z{\mathbb{Z}}

\def\indep{\perp \!\!\! \perp}
\def\notindep{\not\!\perp\!\!\!\perp}

\def\one{\mathbbm{1}}
\usepackage{xcolor}
\definecolor{purple}{RGB}{200,0,255}

\title{DNAMite: Interpretable Calibrated Survival Analysis with Discretized Additive Models}

\author{
       \Name{Mike Van Ness}
       \Email{mvanness@stanford.edu}\\ 
       \addr Stanford University
       \AND
       \Name{Billy Block}
       \Email{biblock@calpoly.edu}\\ 
       \addr California Polytechnic State University
       \AND
       \Name{Madeleine Udell} 
       \Email{udell@stanford.edu}\\ 
       \addr Stanford University\\
}

\begin{document}

\maketitle

\begin{abstract}
Survival analysis is a classic problem in statistics 
with important applications in healthcare.
Most machine learning models for survival analysis are black-box models, limiting their use in healthcare settings where interpretability is paramount. 
More recently, glass-box machine learning models have been introduced for survival analysis, with both strong predictive performance and interpretability.
Still, several gaps remain, as no prior glass-box survival model can produce calibrated shape functions with enough flexibility to capture the complex patterns often found in real data.
To fill this gap, we introduce a new glass-box machine learning model for survival analysis called DNAMite.
DNAMite uses feature discretization and kernel smoothing in its embedding module, making it possible to learn shape functions with a flexible balance of smoothness and jaggedness.
Further, DNAMite produces calibrated shape functions that can be directly interpreted as contributions to the cumulative incidence function.
Our experiments show that DNAMite generates shape functions closer to true shape functions on synthetic data, while making predictions with comparable predictive performance and better calibration than previous glass-box and black-box models.

\end{abstract}

\section{Introduction}

Many healthcare problems require estimating the cumulative distribution function of a time-to-event random variable $T$ given a set of features $X$.
For example, $T$ may represent the time until patient death, and the task is to estimate the mortality rate before time $t$ given patient features $X$ such as demographics, lab measurements, etc.
Such time-to-event problems are often called \emph{survival analysis} problems and have a long history in statistics \citep{clark2003survival}.

While statistical approaches such as the Cox model \citep{cox1972regression} are useful when statistical inference is needed, recent machine learning (ML) approaches for survival analysis often provide better predictive accuracy \citep{ishwaran2008random, lee2018deephit, ren2019deep, hu2021transformer}.
Unfortunately, most of these ML models are black-box, providing little to no explanation for their predictions.
In healthcare settings, model interpretability is critical for several reasons.
First, users (doctors and patients) are more likely to trust a model's predictions if the model can explain how each prediction is generated, including how each of the features contributed to the prediction \citep{poursabzi2021manipulating}.
Second, ML models in healthcare are often trained and validated on data from only one healthcare system, and without model interpretability, spurious signal learned by a model can remain hidden.
For example, visitation by a priest can be highly correlated with mortality \citep{deasy2020dynamic}, and a black-box model could unknowingly rely heavily on this feature despite it potentially not being available in many healthcare systems.
Model interpretability can also enable the discovery of new risk factors or previously unknown patterns in known risk factors, which is not possible with black-box models.
For these reasons, black-box ML models for survival analysis are often greeted with trepidation in healthcare settings.

In contrast to black-box ML models, glass-box ML models are interpretable by design. 
Glass-box ML models can explain how each feature contributes to each prediction (local importance) as well as how each feature affects predictions globally on average (global importance and shape functions).
In the context of supervised learning, recent literature shows that glass-box ML models can achieve performance comparable to black-box models, combining the benefits of black-box ML models and traditional statistical models \citep{nori2019interpretml, popov2019neural, chang2021node, ibrahim2024grand}.
However, few glass-box ML models are available for survival analysis.

\begin{figure*}[t!]
    \centering
    \includegraphics[width=\textwidth]{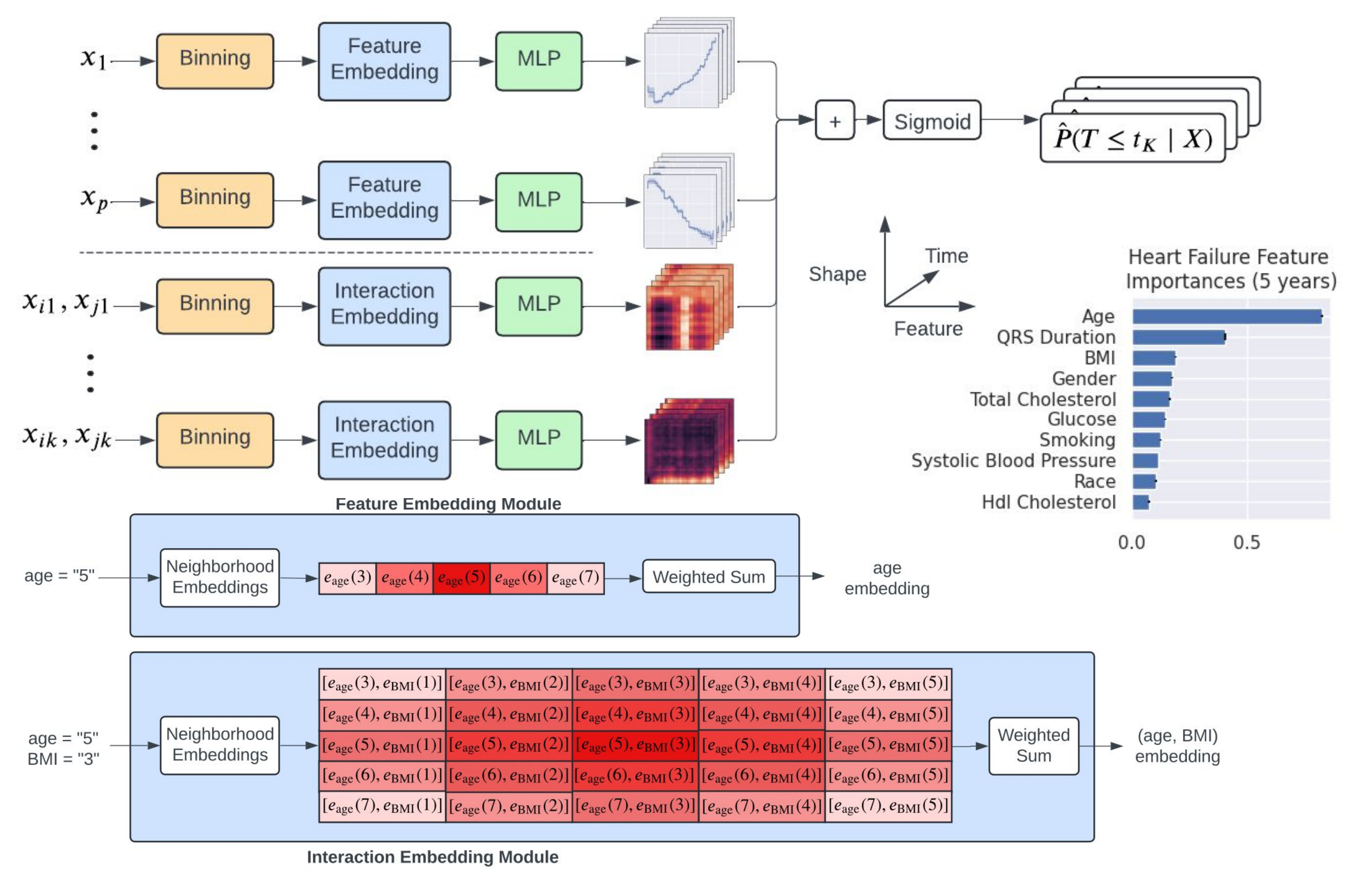}
    \caption{Overview of DNAMite. (Top Left) Each of the $p$ feature and $k$ interaction shape functions consisting of an embedding module followed by a multi-layer perceptron (MLP). The final prediction sums the $K$-dimensional output of each feature/interaction function followed by a sigmoid activation to output $K$ CIF estimates. (Bottom) Main effect and interaction embedding modules. Interaction embeddings concatenate individual feature embeddings. Final embeddings are computed as weighted sums of the embeddings from neighboring feature/interactions. 
    (Right) Example feature importance scores for DNAMite from the heart failure dataset.}
    \label{fig:summary_fig}
\end{figure*}

In this paper, we introduce a new glass-box ML model for survival analysis called the \textit{Discretized Neural Additive Model}, which we refer to as \textbf{DNAMite}.
An overview of DNAMite is given in Figure \ref{fig:summary_fig}.
Following previous work of glass-box models for survival analysis \citep{rahman2021pseudonam, xu2023coxnam, van2024interpretable},
DNAMite is a Neural Additive Model (NAM)
that uses a multi-dimensional prediction head to predict the cumulative distribution function $P(T \leq t \mid X)$ at several evaluation times $t$.
DNAMite offers several key benefits over previous NAMs for survival analysis.
\begin{enumerate}
    \item DNAMite can capture complex signal in each shape function better than previous survival NAMs. 
    While many NAMs learn smooth shape functions, 
    failing to capture more jagged patterns when they appear in real data \citep{agarwal2021neural},
    DNAMite uses feature discretization paired with a specially designed embedding module to balance jaggedness and smoothness in learned feature and interaction shape functions, as demonstrated in Figure \ref{fig:synthetic}. 
    \item DNAMite produces shape functions for each feature/pair that directly describe the contribution of each feature/pair to the distribution function $P(T \leq t \mid X)$, i.e. the most important function to estimate in survival analysis (e.g. probability of death before time $t$ for mortality prediction). In contrast, previous survival NAMs produce shape functions with y-axes related to different quantities, making interpretations less useful. 
    For example, CoxNAM \citep{xu2023coxnam} produces shape functions where the y-axis is the time-independent hazard risk, 
    which does not have a simple clinical interpretation.
    \item DNAMite produces calibrated survival predictions more effectively than previous glass-box survival models. Calibration without post-processing is critical for glass-box models, as post-hoc calibration undermines the interpretability of glass-box models by introducing a nonlinear rescaling between the sum of the feature functions and the final prediction.
\end{enumerate}
Through experiments on synthetic data, we show that DNAMite produces shape functions that more closely match true shape functions.
Further, on real-world survival analysis data, we demonstrate that DNAMite offers predictive performance on par with state-of-the-art glass-box and black-box survival ML models, 
in addition to producing better-calibrated predictions and more useful interpretations.

\section{Background}

\subsection{Survival Analysis}

Given a set of features $X \in \R^p$ and time-to-event label $T \in \R^+$, the survival analysis problem is to estimate the conditional survival probability $P(T > t \mid X)$, or equivalently, the cumulative distribution function $P(T \leq t \mid X)$ which is often called the \emph{cumulative incidence function} (CIF).
To complicate the estimation, $T$ is commonly \emph{censored} for some samples in the training dataset, so that only a lower bound on the time-to-event is known.
Survival training data arrives in the form $(X, Z, \delta)$, where the observed event time $Z = \min(C, T)$ is the minimum of the true event time $T$ and the censoring time $C$, and the censor indicator $\delta = \one_{C > T}$ indicates whether a sample's event time is observed (1) or whether the event is censored (0). 

Classical approaches to survival analysis often model the survival hazard function 
\begin{equation}
    \lambda(t \mid X) = \frac{p(t \mid X)}{P(T > t \mid X)}
\end{equation}
where $p(t \mid X)$ is the conditional density function for $T$. 
For example, the Cox proportional hazards model \citep{cox1972regression} specifies a linear function for the log of the hazard function:
\begin{equation}
    \lambda(t \mid X) = \lambda_o(t) \exp(\beta_1 X_1 + \cdots + \beta_p X_p)
\end{equation}
By separating $\lambda(t \mid X)$ into a time-dependent intercept and a time-independent prediction, the Cox model enforces proportional hazards: for two subjects with features $X$ and $X'$, the ratio $\lambda(t|X) / \lambda(t|X')$ is independent of the time $t$.
Newer ML models relax this assumption either by estimating the hazard function $\lambda(t \mid X)$ (or the CIF/survival function) at several times $t$ \citep{lee2018deephit, ren2019deep, hu2021transformer} or by estimating the parameters of a parametric distribution for $T$ \citep{avati2020countdown}. 

\subsection{Glass-Box Machine Learning}
\label{sec:background}


A glass-box machine learning (ML) model is a model that can produce accurate predictions while being transparent in how predictions are generated. 
Given the ambiguity around model ``interpretability'' \citep{murdoch2019interpretable}, we consider a model to be a glass-box model if 
it can be completely described by a collection of simple plots.
For example, a linear model is a glass-box model since it can be fully described by a single line plot for each feature.
Such glass-box models are crucially different than black-box models pairs with post-hoc interpretability methods like SHAP \citep{lundberg2017unified} or LIME \citep{ribeiro2016should}, as these approaches are inexact and prone to approximation bias \citep{kumar2020problems, van2022tractability}.

Most literature on glass-box ML, including this paper, focuses on generalized additive models (GAMs). Given a set of $p$ features $X = (X_1, X_2, \ldots, X_p)^T$, a GAM $f$ has the form
\begin{equation}
\label{eq:GAM}
    f(X) = \beta_0 + f_1(X_1) + \cdots + f_p(X_p)
\end{equation}
where each feature function $f_j$ is a function of only one feature $X_j$ and often called a \textit{shape function}.
GAMs can also be extended to higher order feature interactions; for example, a GAM with pairwise interactions (called a GA$^2$M) has the form
\begin{equation}
\label{eq:GA2M}
    f(X) = \beta_0 + f_1(X_1) + \cdots + f_p(X_p) + \sum_{j \neq \ell} f_{j, \ell}(X_j, X_\ell).
\end{equation}
In GA$^2$Ms, the individual feature functions are often called \emph{main effects}, while the pairwise interaction functions are called \emph{interaction effects}.

GAMs are glass-box models, since GAMs can be fully described by plotting each shape function.
GAMs can also easily produce feature importances: 
the importance of feature $j$ is $\frac{1}{n} \sum_i |f_j(x_j) |$, which measures how much feature $j$ contributes to the prediction $f(x)$ on average across the training dataset.
Further, GA$^2$Ms are also glass-box models under certain conditions.
First, if the main effects and interaction effects are fit jointly, a purification procedure based on the functional ANOVA decomposition can be used to ``purify'' the main effects so that they maintain the entire marginal signal of each feature \citep{lengerich2020purifying}.
Alternatively, a GA$^2$M can first fit only main effects, freeze the main effects, compute the current prediction residual, and fit the interaction effects on these residuals.

\subsection{Calibration}
\label{sec:calibration_background}

For binary classification problems with binary response $Y \in \{0, 1\}$, a \emph{calibrated} model is a model that produces predictions $\hat{P}(Y = 1 \mid X)$ that correspond in a probabilistic sense to the true value of $P(Y = 1 \mid X)$. 
That is, a model is calibrated if 
for all sets $S = \{i \,:\, \hat{P}(Y^{(i)} = 1 \mid X^{(i)}) \approx \alpha\}$ we have
\begin{equation}
\alpha \approx \hat{P}(Y^{(i)} = 1 \mid X^{(i)})
\approx \sum_{i\in S} \frac{Y^{(i)}}{|S|}.
\end{equation}
For binary classification models, calibration plots can check calibration by comparing prediction quantiles to true positive frequencies \citep{niculescu2005predicting}.
For survival analysis, models are considered calibrated if the CIF $P(T \leq t \mid X)$ is calibrated for all $t$.
Unfortunately, censoring makes it harder to verify calibration, as the true event time $T$ is not known for all users.
The standard approach to check calibration of survival models, which we follow in this paper, is to plot CIF prediction bins against Kaplan-Meier estimates for each bin; see \citep{austin2020graphical} for more details.

\section{Related Work}

While explainable ML methods have a long history in healthcare \citep{abdullah2021review},
methods for explainable survival analysis have only recently been developed \citep{langbein2024interpretable}.
For post-hoc explanations, the SHAP package \citep{lundberg2017unified} supports post-hoc explanations for xgboost models trained with the Cox loss. 
Additionally, recent papers have introduced extensions to SHAP \citep{alabdallah2022survshap, krzyzinski2023survshap} and LIME \citep{kovalev2020survlime} for survival analysis.
For glass-box models, traditional GAMs can be used with the Cox loss \citep{hastie1986generalized}, but other survival losses requires outputting time-dependent prediction, which is made easier with NAMs due to parameter sharing when learning multiple outputs.
As such, multiple NAMs have been proposed for survival analysis with various losses \citep{xu2023coxnam, utkin2022survnam, van2024interpretable, rahman2021pseudonam}.
Nonetheless, none of these previous survival NAMs address the issues of over-smooth shape functions and calibration, which we address with DNAMite.

Even outside of survival analysis, there have been few attempts to mitigate oversmoothing in NAMs.
The original NAMs paper \citep{agarwal2021neural} proposes using exponentially-centered hidden units (ExUs) to learn more jagged shape functions.
However, we find in our experiments that the approach used in DNAMite is much more effective, see Figure \ref{fig:synthetic}.
It is worth noting that traditional GAMs based on splines and decision trees do not over-smooth, 
but are difficult to use for survival analysis since they cannot easily produce multi-dimensional outputs. 

\begin{algorithm}
\caption{DNAMite Training}
\begin{algorithmic}[1]
    \STATE \textbf{Input}: Train data $D = \{(X^{(i)}, Z^{(i)}, \delta^{(i)})\}_{i=1}^n$, \# interactions $k$, \# validation split B.

    \STATE models $\gets [~]$

    \FOR{$B$ iterations}
        \STATE Randomly partition $\{i = 1, 2, \ldots, n\}$  into train samples $\mathcal{T}$ and validation samples $\mathcal{V}$.

        \STATE Train one DNAMite model $f(X) = \sum_j f_j(X_j)$ on $\mathcal{T}$: $ \min_f \frac{1}{|\mathcal{T}|}$, use $\mathcal{V}$ for early stopping.
    
        \STATE Freeze feature functions $f_j, j=1, \ldots, p$.
    
        \STATE Train $f(X) = \sum_j f_j(X_j) + \sum_{j \neq \ell} f_{j, \ell} (X_j, X_\ell)$ on $\mathcal{T}$, use $\mathcal{V}$ for early stopping..

        \STATE Compute intercept for $f$ using Algorithm \ref{alg:intercept} with $f$, $\mathcal{T}$. 

        \STATE models.append($f$).

    \ENDFOR

    \STATE \RETURN models.
    
\end{algorithmic}
\label{alg:dnamite_training}
\end{algorithm}

\section{Methods}

This section describes DNAMite, our proposed glass-box ML model for survival analysis (Figure \ref{fig:summary_fig}).

\subsection{Model Components}

DNAMite is a generalized additive model with pairwise feature interactions. 
DNAMite estimates the CIF $P(T \leq t \mid X)$ by predicting $\hat{P}(T \leq t \mid X)$ at $K$ evaluation times $t = t_1, \ldots, t_K$.
To generate these predictions, DNAMite produces a $K$-dimensional vector for each feature and interaction, which are summed and passed through a sigmoid function to obtain $\hat{P}(T \leq t \mid X)$ for each $t$.
Each feature/interaction function consists of an embedding module followed by a multi-layer perception (MLP).
We use neural networks for each function (instead of splines or boosted decision trees) to facilitate the use of parameter sharing to seamlessly allow each function to have a multivariate output.
The embedding module is imperative to DNAMite, as it is crucial for allowing the model to accurately estimate shape functions (see Figure \ref{fig:synthetic}).
The following two subsections detail the key components of the embedding module.

\subsubsection{Discretization}
\label{sec:discretization}

Unlike previous survival NAMs, DNAMite discretizes both continuous and categorical features.
To discretize a continuous feature into $b$ bins, DNAMite uses feature quantiles to define $b - 1$ cut points that split the data into $b$ unique bins.
One additional bin represents missing values.
After defining the feature bins, DNAMite replaces each original feature value with the corresponding bin index, ranging from $0, 1, \ldots, b$. 
DNAMite learns a different embedding for each possible feature bin.
For categorical features, each unique level gets its own embedding, with one extra bin again for missing values.
For interactions, DNAMite discretizes and embeds each features in the interaction separately and concatenates the two embeddings.

Why discretize a continuous feature? Discretization loses information, yet
counterintuitively, recent evidence suggests this loss of granularity
can improve predictive performance for neural networks \citep{ansari2024chronos, gorishniy2022embeddings, hu2022deepreta}.
One explanation is that discretization, when combined with embedding layers, allows the model to learn embeddings specialized for specific feature ranges, 
which is much more challenging when embedding continuous features without discretization.
Discretization also allows for seamless handling of special feature values.

\subsubsection{Embeddings}
\label{sec:kernel_smoothing}

\begin{figure*}
    \centering
    \includegraphics[width=0.8\linewidth]{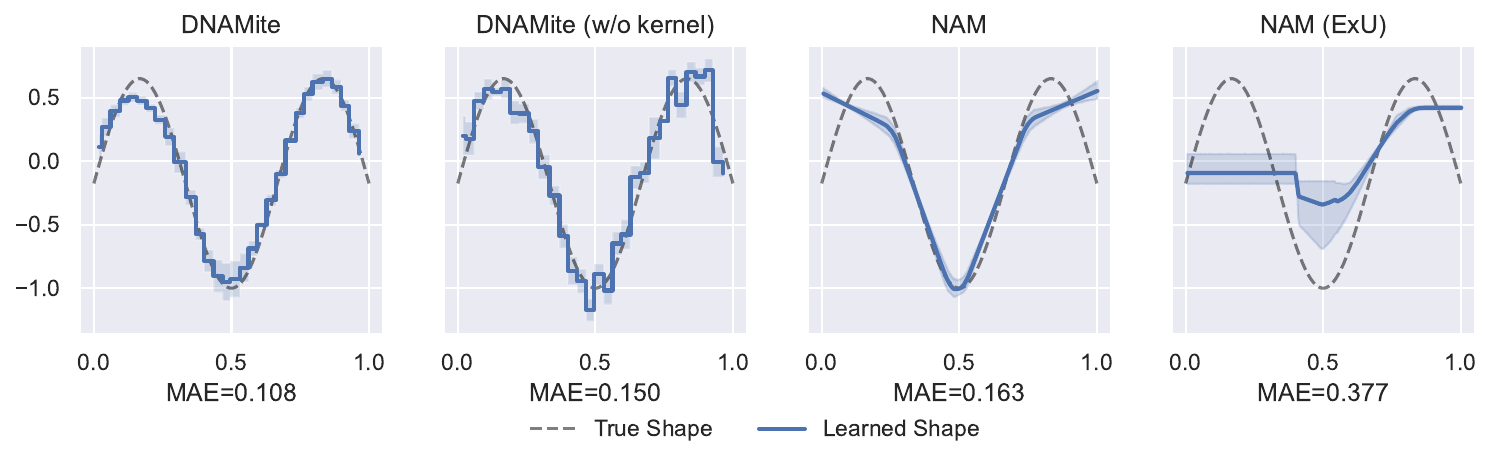}
    \caption{Comparison of model performance in learning one feature's shape function for synthetic data. DNAMite captures the true shape more accurately than competing methods.}
    \label{fig:synthetic}
\end{figure*}

While most NAMs learn overly smooth feature functions, explicit feature discretization causes the opposite problem: learned feature functions can overfit each individual bin, causing jaggedness (as shown in Figures~\ref{fig:synthetic} and \ref{fig:calibration}).
This jaggedness occurs because the embeddings for neighboring feature bins are learned independently, ignoring ordinality.
Such ignorance is different from tree-based models, which also discretize continuous features but make feature splits using ordinal information.

To overcome this limitation, DNAMite learns embeddings for each feature bin that take advantage of the ordering of each continuous feature. 
The embedding module for individual features and interactions is visualized in Figure \ref{fig:summary_fig}.
First, each feature gets a standard embedding module: a lookup table is maintained with $d$-dimensional embeddings for each unique feature bin.
Final embeddings are computed as weighted sums
of the embeddings from neighboring feature bins,
with weights determined by a kernel function that depends on the number of bins between a reference bin and neighbor bin.
A kernel hyperparameter $\gamma$ controls the smoothness of the learned embeddings.
For full details on the embedding module for main effects, see Algorithm \ref{alg:embeds} in Appendix \ref{sec:algorithms}.

\subsection{Interpretability}
\label{sec:interpretability}

DNAMite can produce feature importances and estimated shape functions for the CIF at \emph{each evaluation time} used during model training.
These granular interpretations are critical to capture important patterns in survival data. 
For example, for survival datasets where the proportional hazards assumption is not reasonable, 
time-dependent interpretations allow DNAMite to capture signal that Cox-based models would ignore \citep{van2024interpretable}.
While DyS \citep{van2024interpretable} also produces shape functions for each feature at each time, 
DyS visualizes probability mass estimates $\hat{P}(T = t \mid X)$, which are 
more difficult to interpret than DNAMite's CIF estimates $P(T \leq t \mid X)$.
Importantly, it is not possible to convert DyS's interpretations to CIF-based interpretations because 
the final sigmoid activation is applied to the sum of the features; 
this final nonlinear transformation confounds any additive interpretation via shape functions.
Lastly, PseudoNAM can also output time-depednent interpretations for the CIF, but PseudoNAM involved computing pseudo-values which is very computationally slow for large dataset (too slow to complete in less than 5 hours on 2 out of 5 of our experiment datasets in Table \ref{tab:real_data_benchmark}).

\subsection{Calibration}

As discussed in Section \ref{sec:calibration_background}, glass-box models must be well-calibrated, 
as post-hoc calibration confounds the interpretation of model outputs. 
Unlike previous interpretable survival models, 
DNAMite produces calibrated predictions without post-hoc calibration. 
To produce calibrated predictions, 
DNAMite is trained with the Inverse Probability of Censoring Weighting (IPCW) loss. 
Given CIF predictions $\hat{p}(t_k) = \hat{P}(T \leq t_k \mid X)$, the IPCW loss is given by:
\begin{equation}
    \label{eq:ipcw_loss}
    \sum_{i=1}^n \sum_{k=1}^K \frac{\one_{Z_i > t_k} \hat{p}(t_k)^2}{\hat{P}(C > t_k)} + \frac{\one_{Z_i \leq t_k, \delta_i = 1} (1 - \hat{p}(t_k))^2}{\hat{P}(C > Z_i)}
\end{equation}
The IPCW loss is a proper scoring loss under the assumption that censoring is independent of features, i.e. $C \indep X$ \citep{rindt2022survival}, which means that optimizing the IPCW loss learns asymptotically calibrated predictions.
To estimate the censoring probabilities $\hat{P}(C > t_k)$, we fit a Kaplan-Meier estimator before training DNAMite. 
If there is reason to believe that $C \notindep T$, then as described in \citep{rindt2022survival} the Kaplan-Meier estimate $\hat{P}(C > t_k)$ can be replaced with any calibrated survival estimate $\hat{P}(C > t_k \mid X)$, e.g. a Cox regression model followed by post-hoc calibration \citep{austin2020graphical}.

\begin{table*}[t!]
    \centering
    \caption{Mean Time-Dependent AUC for DNAMite versus baselines. Models are run for 5 trials (mean $\pm$ standard deviation shown), each with a different random seed and train/test split. DNAMite performs similar to the best model on each dataset. Results within one standard deviation of the top performer for each dataset are bold. $\dagger$ indicates a black-box model, and * indicates termination due to high runtime $\geq$ 5 hours.}
    \scalebox{0.9}{
    \begin{tabular}{lccccc}
    \toprule
    dataset &        flchain &       metabric &        support &           unos &  heart\_failure \\
    \midrule
    RSF$^\dagger$     & \textbf{0.955 ± 0.003} & 0.730 ± 0.027 & \textbf{0.831 ± 0.006} & *             & *              \\
    DeepHit$^\dagger$ & \textbf{0.955 ± 0.002} & 0.714 ± 0.022 & 0.752 ± 0.006 & 0.753 ± 0.004 & 0.836 ± 0.004 \\
    SATransformer$^\dagger$ &  \textbf{0.955 ± 0.002} &  0.723 ± 0.026 &  0.826 ± 0.008 &  0.769 ± 0.004 &  \textbf{0.843 ± 0.005} \\
    DRSA$^\dagger$          &  0.939 ± 0.010 &  0.727 ± 0.027 &  \textbf{0.835 ± 0.009} &  0.758 ± 0.004 &  0.809 ± 0.008 \\
    CoxPH         &  0.954 ± 0.002 &  0.707 ± 0.020 &  0.815 ± 0.008 &  0.692 ± 0.003 &  0.827 ± 0.004 \\
    AFT     & 0.954 ± 0.002 & 0.707 ± 0.020 & 0.812 ± 0.008 & 0.689 ± 0.003 & 0.827 ± 0.003  \\
    PseudoNAM    &  \textbf{0.955 ± 0.002} &  0.712 ± 0.016 &  \textbf{0.834 ± 0.008} & * & * \\
    CoxNAM        &  0.928 ± 0.021 &  0.729 ± 0.025 &  0.819 ± 0.008 &  0.752 ± 0.003 &  0.779 ± 0.134 \\
    DyS           &  \textbf{0.957 ± 0.002} &  0.730 ± 0.023 &  \textbf{0.838 ± 0.007} &  0.724 ± 0.003 &  0.829 ± 0.008 \\
    DNAMite (Ours)      &  0.950 ± 0.002 &  \textbf{0.756 ± 0.015} &  \textbf{0.834 ± 0.008} & \textbf{0.779 ± 0.006} &  \textbf{0.841 ± 0.002} \\
    \bottomrule
    \end{tabular}
    }
    \label{tab:real_data_benchmark}
\end{table*}

\begin{figure*}[t!]
    \centering
    \includegraphics[width=0.8\linewidth]{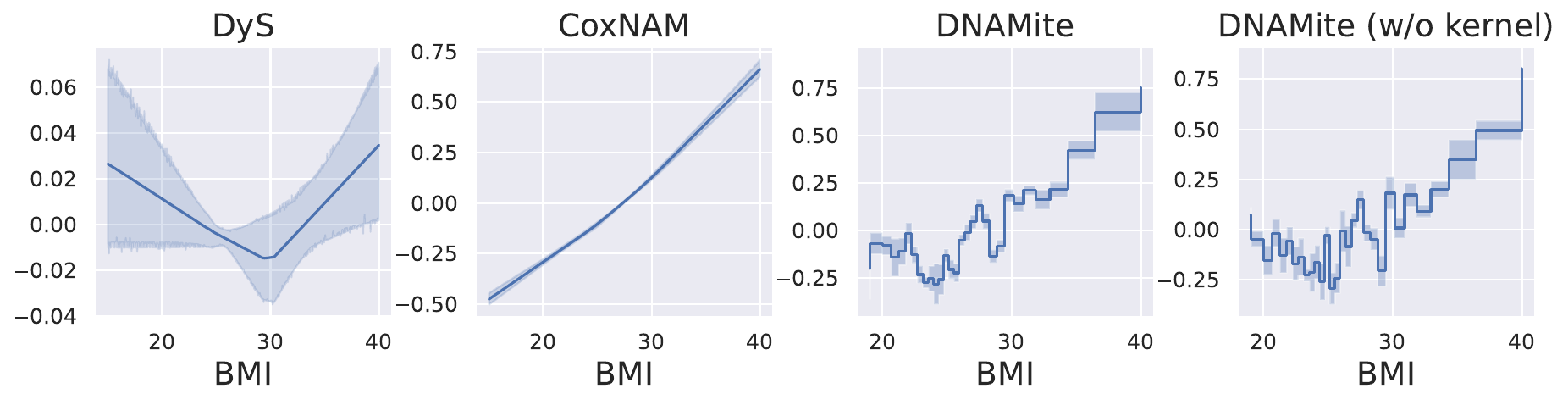}
    \includegraphics[width=\linewidth]{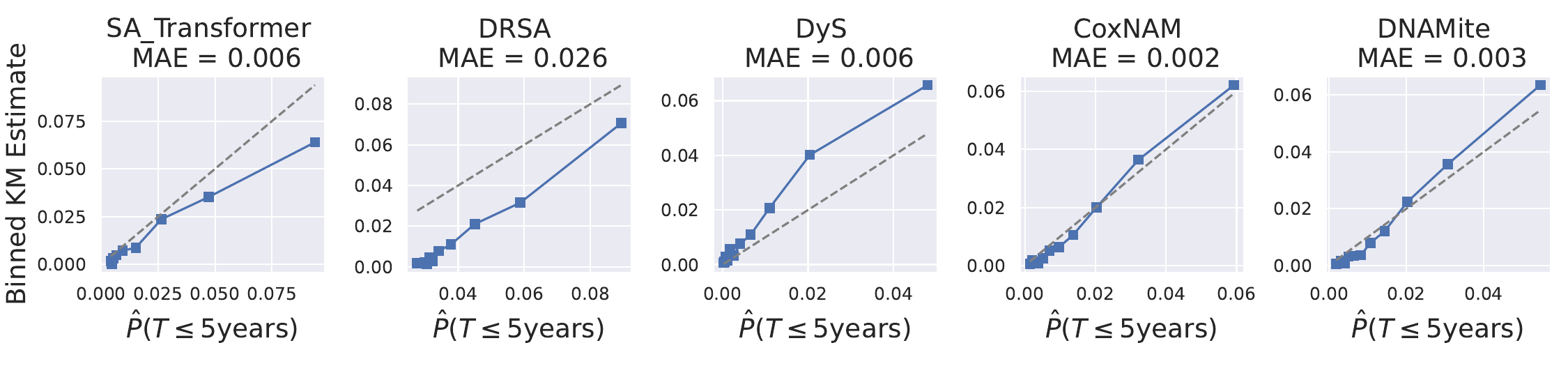}
    \caption{ (Top) Shape functions for BMI feature in heart failure dataset. (Bottom) Calibration plots on heart failure, evaluating at 5 years.}  
    \label{fig:calibration}
\end{figure*}


\subsection{Training Specifications}

The training algorithm for DNAMite is fully described in Algorithm \ref{alg:dnamite_training}.
We highlight a few aspects of the training procedure:

\paragraph{Identification}
GAMs such as DNAMite suffer from unidentifiability without imposing proper constraints \citep{caruana2015intelligible, luber2023structural}. 
For example, for two features $X_j$ and $X_\ell$, shifting $f_j(X_j) \to f_j(X_j) + 1$ and simultaneously shifting $f_\ell(X_\ell) \to f_\ell(X_\ell) - 1$ results in the same predictions but different shape functions.
DNAMite ensures identifiability by constraining the predictions of every shape function
to sum to 0 across the training dataset. 
See Algorithm \ref{alg:intercept} in Appendix \ref{sec:algorithms} for details.

\paragraph{Two-Stage Training} 
DNAMite learns main effects and interaction effects in two stages to avoid the purification step described in Section \ref{sec:background}.
Specifically, main effect terms are first learned via backprop until convergence.
Then the main effect weights are frozen
and the interaction terms are learned through backprop on the prediction function.
As the interaction effects sum embeddings for each of the interacting features,
the (frozen) main effect embeddings are used to warm start the interaction embedding module.

\paragraph{Confidence intervals.}
Confidence intervals for shape functions are critical help diagnose variability in each feature's contribution. 
DNAMite uses cross-validation to generate confidence intervals for shape functions.
DNAmite splits the training data into $B$ different train/validation splits and trains one model on each split. 
The final prediction is the mean prediction from each of the $B$ DNAMite models, and error bars for the shape functions are generated by computing confidence intervals for the mean at each unique bin across the $B$ models.



\section{Experiments}
\label{sec:experiments}

We perform empirical experiments to illustrate the utility of DNAMite compared to existing models. 
Details on our implementation is given in Appendix \ref{sec:implementation}, and 
code for DNAMite can be found at \url{https://github.com/udellgroup/dnamite}.

\subsection{Synthetic Data}

Since true feature shape functions are unknown in real-world data, we generate synthetic survival analysis data to assess the ability of DNAMite and other glass-box models to accurately estimate shape functions.
To evaluate models across both simple and challenging shapes, we generate synthetic survival data where each feature's true shape functions is defined by piece-wise or continuous functions with varying levels of jaggedness. 
Full details on our synthetic data generation can be found in Appendix~\ref{sec:synthetic_appendix}.

We compare DNAMite to three baseline models.
First, to assess the impact of kernel smoothing on the learned shape functions, we set the smoothing parameter in DNAMite to $\gamma = 0$, which we call DNAMite (w/o kernel).
Second, we remove the embedding module entirely from DNAMite and call the resulting model NAM.
Third, we train NAM with ExU activations in an attempt to improve the resulting shape functions \citep{agarwal2021neural}.

Figure~\ref{fig:synthetic} illustrates the performance of DNAMite and other NAMs on one of the synthetic features (additional features are shown in Figure \ref{fig:synthetic_all_feats} in the Appendix). 
We evaluate the ability of each model to capture shape functions by calculating the mean absolute error (MAE) between the predicted and true shape functions.
DNAMite consistently captures feature shapes more accurately, showing lower MAE across all features.
DNAMite handles complex and jagged functions far better than NAM, while maintaining competitive performance on simple feature shapes.
Without kernel smoothing, DNAMite tends to learn overly jagged shape functions, demonstrating the necessity of kernel smoothing for avoiding overfitting.
Surprisingly, despite producing very different shape functions, DNAMite achieves similar predictive performance with and without kernel smoothing.
This result demonstrates that kernel smoothing is necessary to avoid overfitting with respect to true shape functions, but not with respect to predictive performance.
In contrast, NAM learns shape functions that are overly simple and smooth, missing important nuances of feature shapes as a result. 
Additionally, contrary to suggestions from \citep{agarwal2021neural}, we find that using ExU magnifies this issue, resulting in even less accurate shape functions.

\begin{figure*}
    \centering
    \includegraphics[width=\linewidth]{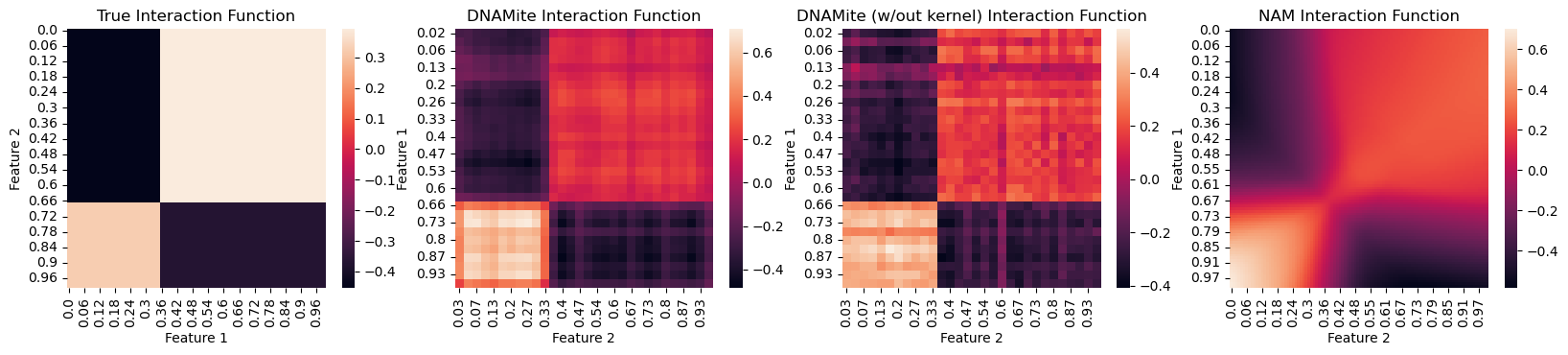}
    \caption{Comparison of models when learning a true shape function on synthetic data.}
    \label{fig:synthetic_interactions}
\end{figure*}

Lastly, we also compare DNAMite to competing methods for learning true interaction functions on synthetic data. 
We generate an interaction term for two features by defining thresholds that split the feature pair into four regions and assigning a different score to each region.
The scores are normalized to have a mean effect of 0 so that interaction effects so not bleed into true main effects.
Figure \ref{fig:synthetic_interactions} shows that DNAMite can more accurately learn interactions compared to other models.

\subsection{Real Data Benchmark}

\begin{table}[]
    \centering
    \caption{DNAMite performance on various real-world datasets as a function of the maximum number of bins used during feature discretization. }
    \scalebox{0.9}{
    \begin{tabular}{lccc}
    \toprule
        bins & support       & unos          & heart\_failure \\
        \midrule
         8    & 0.830 ± 0.008 & 0.768 ± 0.004 & 0.829 ± 0.003  \\
        16   & 0.834 ± 0.007 & 0.773 ± 0.005 & 0.841 ± 0.003  \\
        32   & 0.831 ± 0.008 & 0.777 ± 0.005 & 0.842 ± 0.002  \\
        64   & 0.824 ± 0.008 & 0.778 ± 0.004 & 0.839 ± 0.002 \\
        \bottomrule
    \end{tabular}
    }
    \label{tab:max_bins}
\end{table}

\begin{table}
\centering
\caption{Run times (wallclock seconds) for unos and heart\_failure. Times for the other datasets are given in Table \ref{tab:runtimes_extra} in the Appendix.}
\scalebox{0.8}{
\begin{tabular}{lcc}
\toprule
model          & unos               & heart\_failure     \\ 
\midrule
CoxPH         & 22.895 ± 1.257     & 17.920 ± 2.906     \\ 
AFT            & 12.073 ± 0.282     & 9.378 ± 0.201      \\ 
SATransformer & 5414.365 ± 602.512 & 2100.597 ± 95.863  \\ 
DRSA           & 1157.576 ± 76.178  & 360.696 ± 26.253   \\ 
RSF            & *                  & *                  \\ 
CoxNAM        & 657.999 ± 89.798   & 165.809 ± 31.627   \\ 
PseudoNAM      & *                  & *                  \\ 
DyS            & 1517.205 ± 86.914  & 783.335 ± 165.016  \\ 
DNAMite        & 2690.676 ± 103.465 & 819.684 ± 78.972   \\ 
\bottomrule
\end{tabular}
}
\label{tab:runtimes}
\end{table}


We now evaluate DNAMite on real-world survival data, comparing to the following other glass-box and black-box ML models. Details on hyperparameters can be found in Appendix \ref{sec:hyperparameters}, and additional training details can be found in Appendix \ref{sec:training_details}.
\begin{itemize}
    \item \textbf{CoxPH}: linear Cox model \citep{cox1972regression}.
    \item \textbf{AFT}: Accelerated Failure Time model, another linear model \citep{wei1992accelerated}. 
    \item \textbf{RSF}: Random Survival Forest, a black-box tree-based survival model \citep{ishwaran2008random}.
    \item \textbf{DeepHit}: a black-box model using neural networks \citep{lee2018deephit}.
    \item \textbf{DRSA}: a black-box model which uses recurrent neural networks to output survival predictions at multiple evaluation times \citep{ren2019deep}.
    \item \textbf{SATransformer}: a black-box model based on transformers \citep{hu2021transformer}.
    \item \textbf{CoxNAM}: glass-box model using Cox loss \citep{xu2023coxnam}.
    \item \textbf{PseudoNAM}: glass-box model trained using pseudo-values for survival curve \citep{rahman2021pseudonam}.
    \item \textbf{DyS}: glass-box model using a discrete-time model and the RPS loss \citep{van2024interpretable}.
\end{itemize}

The time-dependent AUC (as defined in scikit-survival \citep{polsterl2020scikit}) for each dataset and model are shown in Table \ref{tab:real_data_benchmark}.
Full dataset details can be found in Table \ref{tab:openml_datasets} in the appendix.
We use time-dependent AUC for model evaluation since all models (except CoxPH and CoxNAM) produce time-dependent predictions. 
While C-index is generally not suitable for comparing survival models with time-dependent predictions, we also report time-independent C-index scores for all models in Table \ref{tab:cindex} in the appendix for completeness.
On two datasets (metabric and unos), DNAMite is the best model, and there is only one dataset (flchain) where DNAMite is not within one standard deviation from the best model, including black-box models.
This demonstrates that DNAMite is competitive with state-of-the-art ML models for survival analysis, despite being a glass-box model.
Further, DNAMite is the best performing glass-box model on 3 out of 5 datasets, showcasing DNAMite as a state-of-the-art glass-box survival model.

\subsubsection{Shape Functions}

The top portion of Figure \ref{fig:calibration} shows shape functions for the BMI feature in the heart failure data, using an evaluation time of 5 years. 
In addition to the glass-box baseline models, we include a version of DNAMite setting $\gamma = 0$ in the embedding module to assess the impact of the kernel smoothing on the learned shape functions, as done in Figure \ref{fig:synthetic}.
The results are similar to the synthetic data results: both DyS and CoxNAM produce shape functions that are too smooth, while DNAMite without smoothing produces a shape function that is too jagged. 
Meanwhile, DNAMite's shape function is non-monotonic and captures details of the BMI feature without being too noisy.

DNAMite's shape function also has the most direct interpretation.
In the context of the heart failure dataset, DNAMite's shape function can be interpreted as the contribution of BMI to the predicted probability of developing heart failure in the next 5 years (on the log-odds scale).
DyS's shape function, meanwhile, represents the contribution of BMI to the predicted probability of developing heart failure \emph{at} 5 years (not before or after, pre-softmax), which is less useful.
CoxNAM's shape function represents the contribution of BMI to the time-independent hazard risk of heart failure, which has no simple clinical interpretation. 

\subsubsection{Calibration}

The bottom portion of Figure \ref{fig:calibration} shows calibration plots for the competing methods using the plotting method described in Section \ref{sec:calibration_background}, along with the MAE of each plot from the optimal calibration line (shown with dotted line).
Calibration plots for additional datasets are shown in Figure~\ref{fig:extra_calibration_plots}.
DNAMite and CoxNAM are both well calibrated, while other models have worse calibration visually and in terms of calibration MAE.
Additionally, Table~\ref{tab:brier_scores} in the appendix shows brier scores for each model, which further demonstrates that DNAMite is the most calibrated model.

\subsubsection{Runtimes}
Table \ref{tab:runtimes} shows the runtimes for all models considered in Table \ref{tab:real_data_benchmark}. All DNAMite runtimes are less than 1 hour, positioning DNAMite as a relatively efficient option even for larger datasets. To achieve these runtimes it is critical to use the implementation tricks discussed in Appendix \ref{sec:implementation}.

\section{Conclusion}

We present DNAMite, a new glass-box ML model for survival analysis based on neural additive models.
DNAMite fills several holes in existing literature for interpretable survival analysis.
Notably, DNAMite uses an embedding module that combines feature discretization and kernel smoothing, resulting in shape functions that can more accurately capture complex feature signal.
Additionally, DNAMite's shape functions are well-calibrated and convey contributions directly to the cumulative incidence function, leading to easier and more useful clinical interpretations.
In healthcare settings where predictive accuracy, calibration, and interpretability are all paramount, DNAMite represents a step forward for modeling of clinical survival problems that we hope will inspire further research in interpretable survival analysis.

\bibliography{references} 

\appendix

\section{Algorithms}
\label{sec:algorithms}

\begin{algorithm}[H]
\caption{Feature Embeddings}
\begin{algorithmic}[1]
    \STATE \textbf{Input}: feature value $x \in \R$, binning function $b : \R \to \Z^+$, embedding function $e : \Z^+ \to \R^d$\\ kernel strength $\gamma$, kernel width $k$, max bin size $B$.

    \STATE Replace $x$ with bin index $b(x)$.

    \STATE $\text{kernel\_weights} \gets [\exp\left(-\frac{(x - z)^2}{2\gamma}\right) \text{ for } z \in [-k, k] \cap \Z]$

    \STATE $\text{neighbor\_bins} \gets [x + z \text{ for } z \in [-k, k] \cap \Z]$.

    \STATE Remove neighbors where $\text{neighbor\_bins}_i \leq 0$ or $\text{neighbor\_bins}_i > B$.

    \STATE $\text{embeds} \gets [e(\text{index}) \text{ for index in neighbor\_bins}]$ 

    \STATE \RETURN $\sum_i \text{kernel\_weights}_i \cdot \text{embeds}_i$.
    
\end{algorithmic}
\label{alg:embeds}
\end{algorithm}

\begin{algorithm}[H]
\caption{Compute Intercept}
\begin{algorithmic}[1]
    \STATE \textbf{Input}: DNAMite model $f$, train data $D = \{(X^{(i)}, Y^{(i)})\}_{i=1}^n$.

    \STATE $\beta_0 \gets 0$.

    \FOR{$j = 1, \ldots p$}

        \STATE $f_j \gets f_j - \frac{1}{n} \sum_{i=1}^n f_j (X_j^{(i)})$

        \STATE $\beta_0 \gets \beta_0 + \frac{1}{n} \sum_{i=1}^n f_j (X_j^{(i)})$
        
    \ENDFOR

    \FOR{pairs $j, \ell$ in $f$}

        \STATE $f_{j, \ell} \gets f_{j, \ell} - \frac{1}{n} \sum_{i=1}^n f_{j, \ell} (X_j^{(i)}, X_\ell^{(i)})$

        \STATE  $\beta_0 \gets \beta_0 + \frac{1}{n} \sum_{i=1}^n f_{j, \ell} (X_j^{(i)}, X_\ell^{(i)})$
    
    \ENDFOR

    \STATE Add $\beta_0$ to $f$

    \STATE \RETURN $f$.
    
\end{algorithmic}
\label{alg:intercept}
\end{algorithm}

\section{DNAMite Implementation}
\label{sec:implementation}

NAMs must be implemented carefully in order to not suffer from inefficiency. 
The main bottleneck in most NAM implementations is an explicit PyTorch for loop to do the forward pass through each individual feature/interaction function.
In order to avoid using an explicit for loop, we use the following implementation tricks to create an efficient implementation for DNAMite.

\paragraph{Stacking embeddings}
In applications such as natural language processing, one embedding layer is used to embed all features, e.g. all tokens in a text sequence.
In tabular prediction, however, each feature gets its own embedding, since the bin indices of one feature do not represent the same quantity as the bin indices of the other features.
The naive implementation, then, is to create one torch embedding layer for each feature.
However, this requires a manual for loop, which is exactly what we are trying to avoid.
Thus, an alternative solution is to stack all of the individual feature embeddings into one ``super'' PyTorch embedding layer.
In order to implement this, for each feature $X_j$ we can store an embedding offset to indicate the first index in the super embedding that corresponds to $X_j$. To be explicit, suppose that feature $X_j$ has 10 unique bins, and the offset for $X_j$ was 100, then indices $100, 101, \ldots, 109$ would correspond to the 10 unique $X_j$ bins in the PyTorch embedding layer.

\paragraph{Using einsum}
DNAMite uses an MLP for each feature/interaction function, which means that naively we would store a PyTorch ModuleList of MLPs, one for each feature/interaction.
Suppose each feature $X_j$ has an MLP with a linear layer weight $W_j$ with shape $(d, d)$ for some hidden dimension $d$, which multiplies a batched input $H_j$ with shape $(N, d)$ to produce $Y_j = H_j W_j$. 
To combine these layers into one layer, we can stack the linear layer weights into one weight $W$ with shape $(p, h, h)$, where $p$ is the number of features.
Then we can compute $Y_j$ for each $j = 1, \ldots, p$ simultaneously using the einsum call
\[
\text{torch.einsum}('ijk,jkl\to jkl', H, W)
\]
For complete implementation details, please refer to our code \url{https://github.com/udellgroup/dnamite}.

\section{Hyperparameters}
\label{sec:hyperparameters}

There are 3 important hyperparameters that need to be set in DNAMite.
\begin{enumerate}
    \item \textbf{Kernel smoothing parameter $\gamma$ }: setting $\gamma = 0$ decreases smoothness too much as shown in Figures \ref{fig:synthetic} and \ref{fig:calibration}, while using $1 \geq \gamma \geq 3$ yields similar smoothness in our experience. 
    Further, different values of $\gamma$ often yield similar accuracy, so we recommend setting $\gamma$ to achieve the desired smoothness, especially when true shape functions are unknown for non-synthetic data.
    \item \textbf{Max bins}: Table \ref{tab:max_bins} shows the impact of changing max bins. Since 32 works well across datasets, we generally recommend using this value. 
    \item \textbf{Hidden dimension}: we’ve found that using a hidden dimension $\geq 32$ works well across datasets. Using a smaller hidden dim (e.g. 8) sometimes also works well on smaller datasets.
\end{enumerate}

\section{Datasets}

\begin{table}[h]
    \centering
    \caption{Datasets used in our experiments. For datasets coming from OpenML, the OpenML ID is listed for ease of reproducibility.}
    \label{tab:openml_datasets}
    \vspace{0.5pc}
    \begin{tabular}{lccc}
        \toprule
        Name & n & p & Censor Rate\\
        \midrule
        metabric & 1981 & 79 & 0.552 \\
        flchain & 7874 & 9 & 0.725 \\
        support & 9105 & 42 & 0.319 \\
        heart failure & 265818 & 3108 & 0.974 \\
        unos & 574819 & 20 & 0.907 \\
        \bottomrule
    \end{tabular}
\end{table}

Table \ref{tab:openml_datasets} lists the datasets used in our experiments.
We provide additional details for the origin of each of these datasets.
\begin{itemize}
    \item \textbf{heart failure}: we obtain this dataset from the electronic health records from a large hospital network (name censored for anonymity). 
    Features are taken from the PCP-HF model for heart failure \citep{khan201910}.
    The survival event is time until a patient gets heart failure.
    \item \textbf{unos}: this data comes from the UNOS transplant data, which is publicly accessible via request at \url{https://unos.org/data/}.
    We get data from the ``kidpan'' dataset, which contains data on kidney and pancreas transplants. 
    We use data from non-multi-organ patients on the kidney waitlist after January 1, 2010.
    The survival label is the time until a patient on the transplant waitlist dies.
    \item \textbf{support}: dataset to predict survival time from death for critically ill patients.
    We obtain the data from the SurvML python package: \url{https://github.com/survml/survml-deepsurv/blob/main/data/support_parsed.csv}.
    \item \textbf{flchain}: we obtain via scikit-survival \citep{polsterl2020scikit}.
    \item \textbf{metabric}: we obtain via the GitHub repository for the DeepHit paper  \footnote{\url{https://github.com/chl8856/DeepHit/tree/master/sample\%20data/METABRIC}} \citep{lee2018deephit}.
\end{itemize}

\section{Additional Experiment Details}
\label{sec:training_details}

For all deep learning models, we use the Adam optimizer \citep{kingma2014adam} with a learning rate of 0.0005, we which we found to work well across all models and datasets.
We use a batch size of 64 for metabric, 128 for flchain and support, and 512 for heart failure and unos, reflective the relative size of each dataset.
All deep learning models are trained using a single GPU.
We run each deep learning model for a maximum of 100 epochs.
In order to avoid overfitting, we early stop a model if the validation loss does not decrease for 5 straight epochs, and revert the model's weights to the epoch with the best validation loss.
All DNAMite models use $\gamma = 1$ as part of the embedding module, unless explicitly stated that kernel smoothing is not used, in which case $\gamma = 0$.

\subsection{Synthetic Data}
\label{sec:synthetic_appendix}

\begin{figure*}
    \centering
    \includegraphics[width=\linewidth]{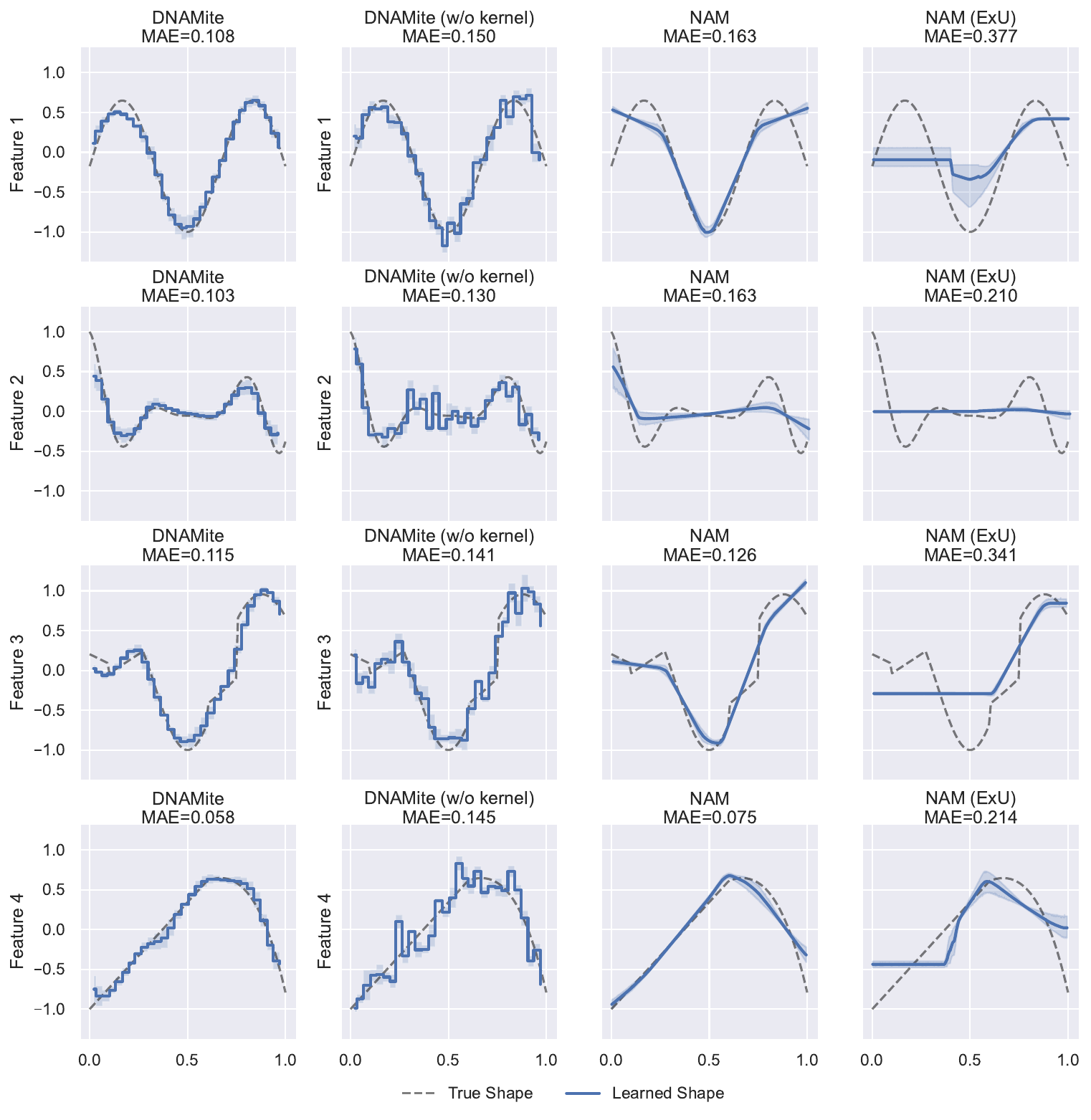}
    \caption{Shape functions for all synthetic features to complement the results in Figure~\ref{fig:synthetic}. DNAMite consistently exhibitis the lowest MAE across all features.}
    \label{fig:synthetic_all_feats}
\end{figure*}

\paragraph{Data Generation}
We independently sample each feature from a $\text{Uniform}(0,1)$ distribution, restricting our analysis to the interval $[0, 1]$.
We calculate risk as the sum of the output of the feature functions, plus a small amount of Guassian noise.
This risk score is then transformed using a sigmoid function to represent the probability $p$ of the event occurring before some evaluation time $t$.
To generate the true response $T$, we draw a Bernoulli random variable $X$ with parameter $p$. If $X = 1$ we draw $T$ from a $\text{Uniform}(0, t)$ distribution, and if $X = 0$ we draw $T$ from a $\text{Uniform}(t, t_{\text{max}})$ distribution.

\paragraph{Univariate Feature Shape Functions}
We define four true univariate feature functions with varying degrees of complexity:
\begin{enumerate}

    \item Sine function with period $T=3/pi$:
    \[
    f_1(x) = \sin(3 \pi x)
    \]
    \item Complex cosine and polynomial function:
    \[
    f_2(x) = \left(\cos(5.2 \pi x) + 0.5x\right) (x - 0.5)^2
    \]
    \item Piecewise function with five segments, including sine/cosine functions and polynomials, and three discontinuities:
    \[
    f_3(x) =
    \begin{cases} 
    -\frac{3}{2}x + 1 & \text{if } x < 0.1 \\
    2x & \text{if } 0.1 \leq x < 0.275 \\
    \sin(3 \pi x) & \text{if } 0.275 \leq x < 0.6 \\
    2x^2 - 1 & \text{if } 0.6 \leq x < 0.75 \\
    \cos(2.3 \pi x) + 0.5x & \text{if } x \geq 0.75
    \end{cases}
    \]
    \item Simple piecewise function with no discontinuities and two segments: a straight line and a polynomial:
    \[
    f(x) =
    \begin{cases} 
    x & \text{if } x < 0.6 \\
    -x^5 + x + (0.6)^5 & \text{if } x \geq 0.6
    \end{cases}
    \]
    
\end{enumerate}
Each feature function is normalized to have an average value of 0 and scaled such that all values lie in the range $[-1, 1]$.

\paragraph{Interaction Shape Functions}
We generate true interaction term shape functions by defining thresholds \( p_1 \) and \( p_2 \) for features \( x_1 \) and \( x_2 \) that segment the interaction:
\[
f(x_1, x_2) =
\begin{cases}
    w_{00} & \text{if } x_1 < p_1 \text{ and } x_2 < p_2, \\
    w_{01} & \text{if } x_1 < p_1 \text{ and } x_2 \geq p_2, \\
    w_{10} & \text{if } x_1 \geq p_1 \text{ and } x_2 < p_2, \\
    w_{11} & \text{if } x_1 \geq p_1 \text{ and } x_2 \geq p_2,
\end{cases}
\]
\noindent where \( w_{00}, w_{01}, w_{10}, \) and \( w_{11} \) are weights assigned to each region. We center these weights around zero by subtracting the area-weighted mean to prevent interaction effects from bleeding into main effects.

We define one interaction term between features 1 and 2 in our synthetic data generation, visualized in Figure \ref{fig:synthetic_interactions}.

\begin{figure*}
    \centering
    \includegraphics[width=0.8\linewidth]{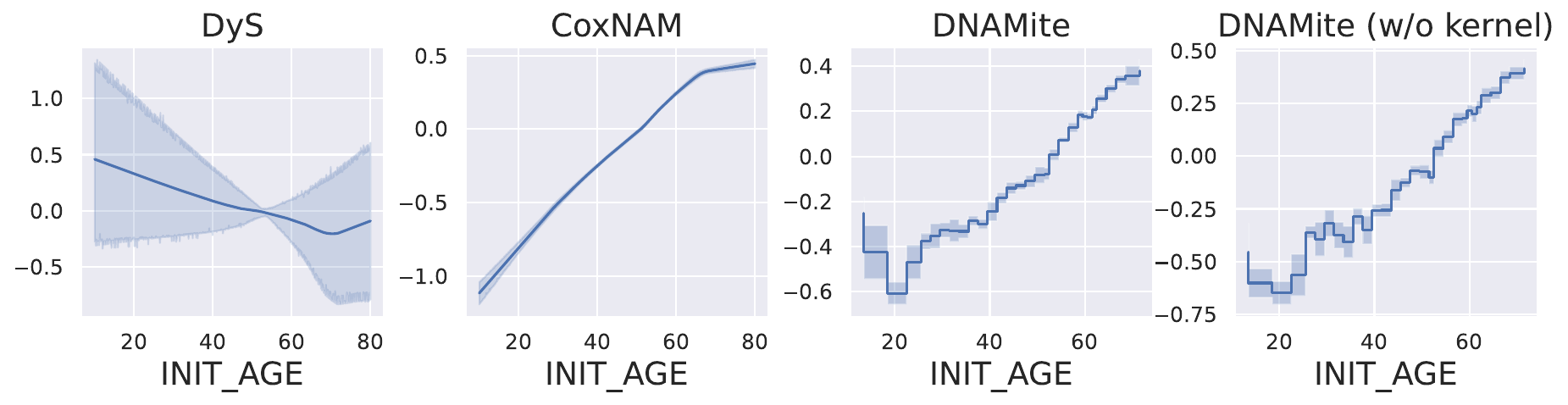}
    \includegraphics[width=\linewidth]{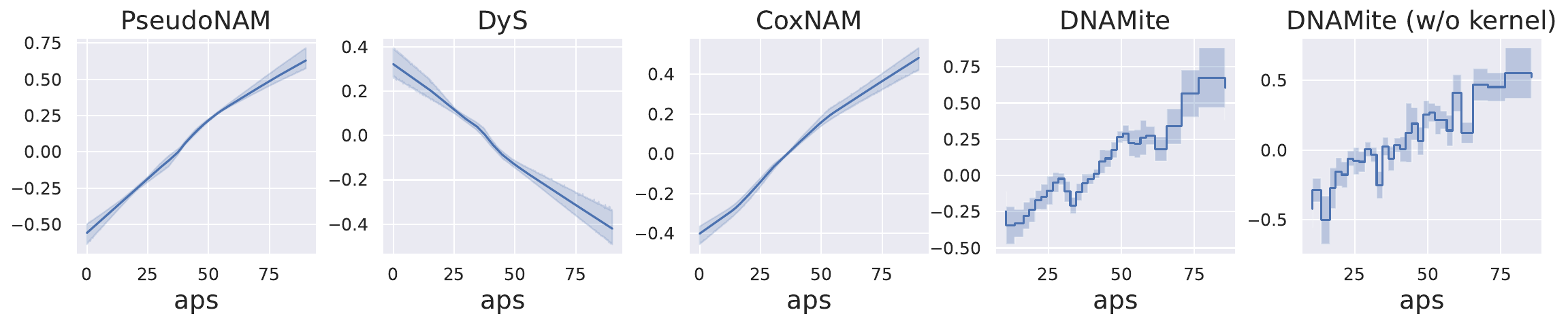}
    \includegraphics[width=\linewidth]{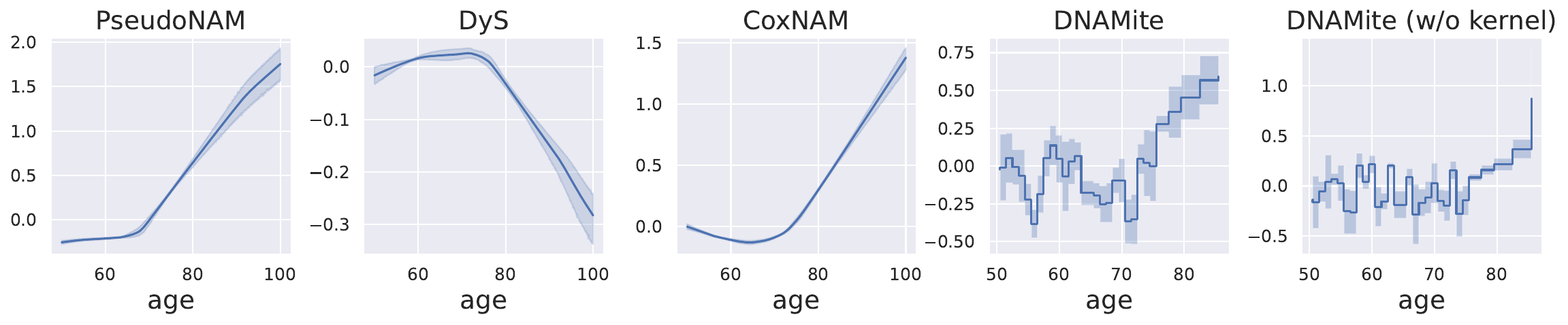}
    \includegraphics[width=\linewidth]{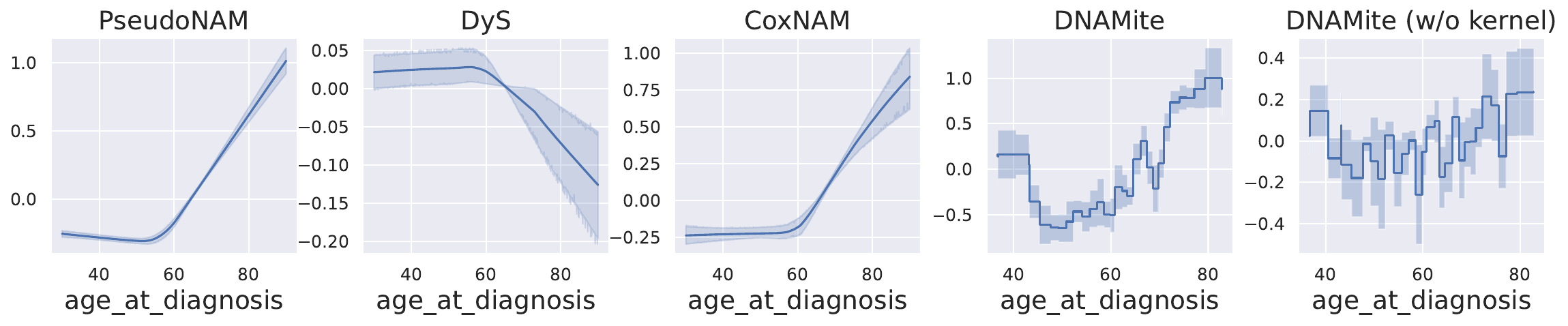}
    \caption{Shape functions for competing models to complement the results in Figure \ref{fig:calibration}. Rows in order: 1) INIT\_AGE feature from unos (2 years) 2) aps feature from support (2 years) 3) age feature from flchain (10 years) 4) age at diagnosis feature from metabric (10 years).}
    \label{fig:extra_shape_functions}
\end{figure*}

\begin{table*}[]
    \centering
    \caption{Brier scores to complement the results in Table \ref{tab:real_data_benchmark}.}
    \begin{tabular}{lccccc}
    \toprule
    dataset &        flchain &       metabric &        support &           unos &  heart\_failure \\
    \midrule
    CoxPH         &  0.045 ± 0.000 &  \textbf{0.179 ± 0.013} &  \textbf{0.173 ± 0.004} &  0.128 ± 0.002 &  0.024 ± 0.000 \\
    SATransformer &  0.045 ± 0.000 &  0.194 ± 0.015 &  0.176 ± 0.005 &  0.159 ± 0.005 &  0.027 ± 0.000 \\
    DRSA          &  0.061 ± 0.002 &  0.207 ± 0.024 &  0.194 ± 0.007 &  0.123 ± 0.001 &  0.024 ± 0.000 \\
    CoxNAM        &  0.072 ± 0.024 &  \textbf{0.175 ± 0.015} &  \textbf{0.170 ± 0.004} &  0.118 ± 0.002 &  \textbf{0.024 ± 0.001} \\
    PseudoNAM     &  0.060 ± 0.002 &  0.265 ± 0.024 &  \textbf{0.176 ± 0.008} &      nan ± nan &      nan ± nan \\
    DyS           &  \textbf{0.044 ± 0.001} &  0.203 ± 0.029 &  0.179 ± 0.007 &  0.156 ± 0.001 &  0.027 ± 0.001 \\
    DNAMite       &  0.046 ± 0.000 &  \textbf{0.166 ± 0.010} &  \textbf{0.169 ± 0.006} &  \textbf{0.111 ± 0.002} &  \textbf{0.023 ± 0.000} \\
    \bottomrule
    \end{tabular}
    \label{tab:brier_scores}
\end{table*}

\begin{table*}[]
    \centering
    \caption{C-index scores to complement the results in Table \ref{tab:real_data_benchmark}. For all models that produce time-dependent predictions, we use the median evaluation time as the risk score for computing C-index.}
    \begin{tabular}{lccccc}
        \toprule
        dataset &        flchain &       metabric &        support &           unos &  heart\_failure \\
        \midrule
        CoxPH         & 0.933 ± 0.001 & 0.662 ± 0.006 & 0.735 ± 0.006 & 0.678 ± 0.001 & 0.840 ± 0.003  \\
        AFT           & 0.933 ± 0.001 & 0.662 ± 0.007 & 0.733 ± 0.006 & 0.678 ± 0.001 & 0.841 ± 0.003  \\
        SATransformer & 0.932 ± 0.002 & 0.663 ± 0.014 & 0.738 ± 0.007 & 0.742 ± 0.003 & 0.852 ± 0.004  \\
        DRSA          & 0.934 ± 0.001 & 0.672 ± 0.007 & 0.738 ± 0.007 & 0.724 ± 0.004 & 0.843 ± 0.006  \\
        RSF           & 0.929 ± 0.002 & 0.702 ± 0.014 & 0.733 ± 0.004 & *             & *              \\
        CoxNAM        & 0.936 ± 0.001 & 0.677 ± 0.008 & 0.741 ± 0.006 & 0.736 ± 0.002 & 0.852 ± 0.004  \\
        PseudoNAM     & 0.932 ± 0.003 & 0.662 ± 0.030 & 0.741 ± 0.006 & *             & *              \\
        DyS           & 0.935 ± 0.001 & 0.672 ± 0.007 & 0.744 ± 0.006 & 0.719 ± 0.007 & 0.847 ± 0.006  \\
        DNAMite       & 0.928 ± 0.002 & 0.673 ± 0.011 & 0.737 ± 0.007 & 0.734 ± 0.004 & 0.851 ± 0.002 \\
        \bottomrule
    \end{tabular}
    \label{tab:cindex}
\end{table*}

\begin{table*}
\centering
\caption{Run times (wallclock seconds) for remaining models to accompany Table \ref{tab:runtimes}.}
\begin{tabular}{lccc}
\toprule
model          & flchain         & metabric          & support          \\ 
\midrule
CoxPH         & 1.939 ± 0.070   & 0.719 ± 0.031     & 0.641 ± 0.102    \\ 
AFT            & 0.994 ± 0.049   & 0.690 ± 0.030     & 1.282 ± 0.182    \\ 
SATransformer & 140.570 ± 7.604 & 51.691 ± 4.782   & 109.877 ± 5.244  \\ 
DRSA           & 32.416 ± 4.121  & 12.760 ± 1.092    & 23.909 ± 1.168   \\ 
RSF            & 11.550 ± 0.271  & 2.978 ± 0.095     & 32.924 ± 0.181   \\ 
CoxNAM        & 36.808 ± 4.285  & 29.347 ± 4.361    & 28.774 ± 4.018   \\ 
PseudoNAM      & 147.307 ± 2.560 & 49.969 ± 8.032    & 161.828 ± 4.217  \\ 
DyS            & 114.990 ± 9.820 & 115.324 ± 13.648  & 155.834 ± 30.686 \\ 
DNAMite        & 60.836 ± 10.967 & 365.043 ± 104.324 & 165.845 ± 12.789 \\ 
\bottomrule
\end{tabular}
\label{tab:runtimes_extra}
\end{table*}

\begin{figure*}
    \centering
    \includegraphics[width=\linewidth]{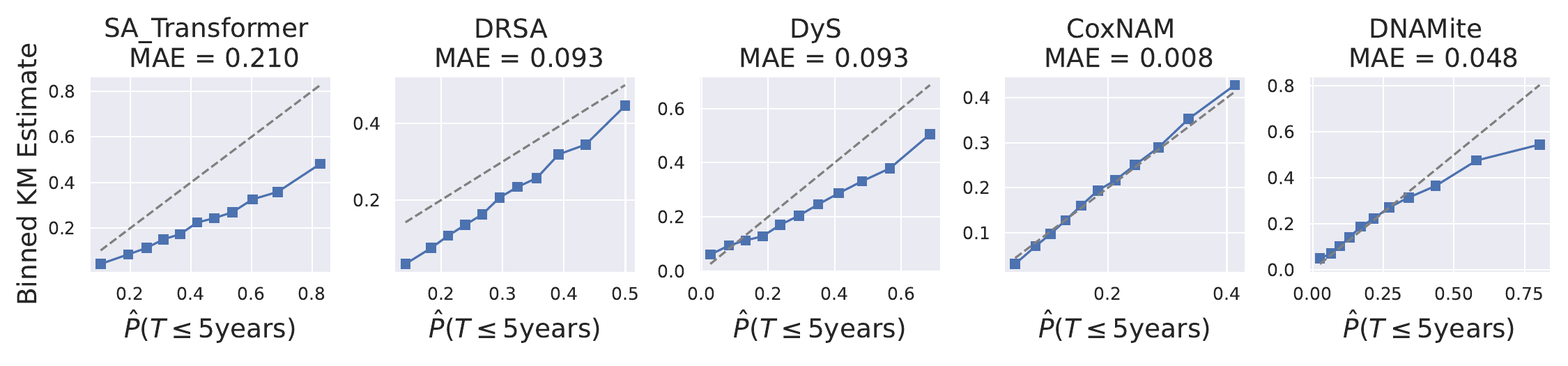}
    \includegraphics[width=\linewidth]{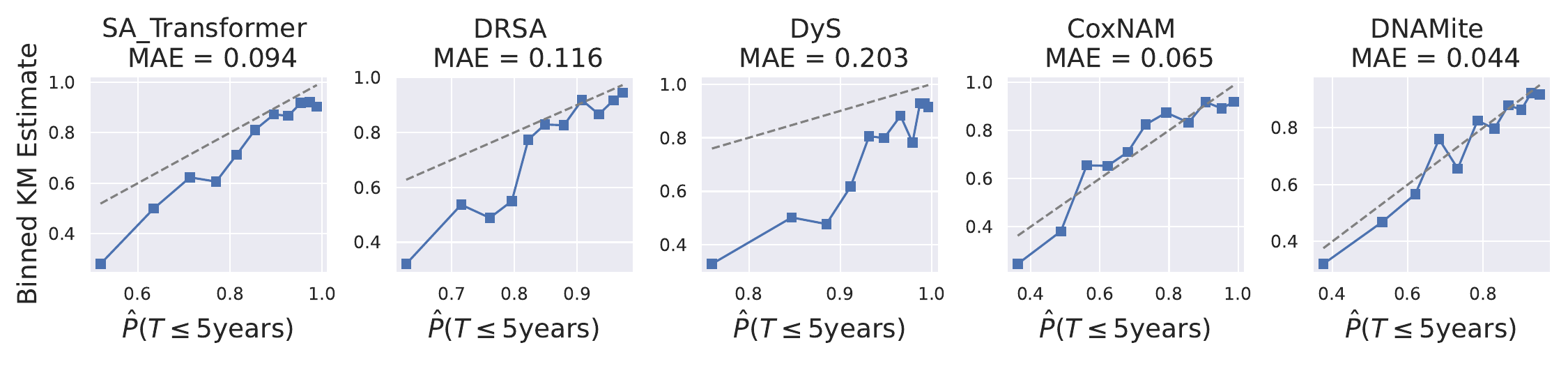}
    \includegraphics[width=\linewidth]{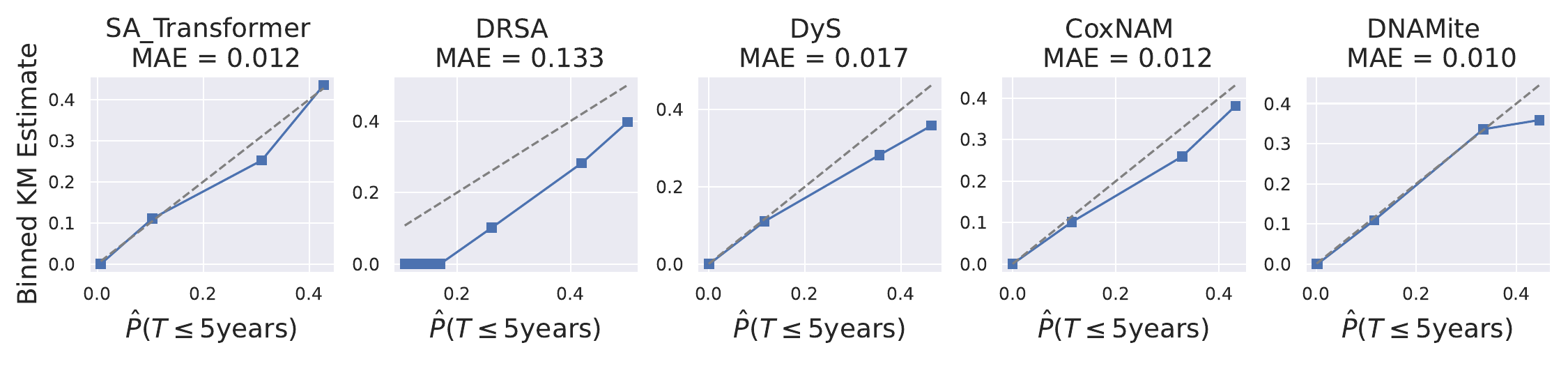}
    \includegraphics[width=\linewidth]{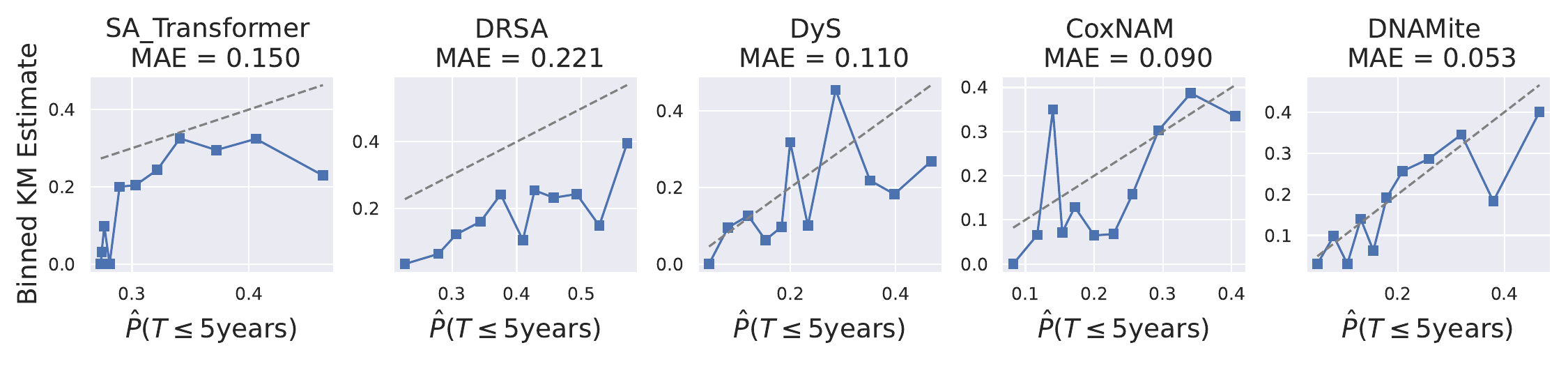}
    \caption{Calibration plots for other datasets in benchmark to compliment the results in Figure \ref{fig:calibration}. Rows in order: 1) unos (2 years) 2) support (2 years) 3) flchain (10 years) 4) metabric (10 years).}
    \label{fig:extra_calibration_plots}
\end{figure*}

\begin{figure*}
    \centering
     \includegraphics[width=\linewidth]{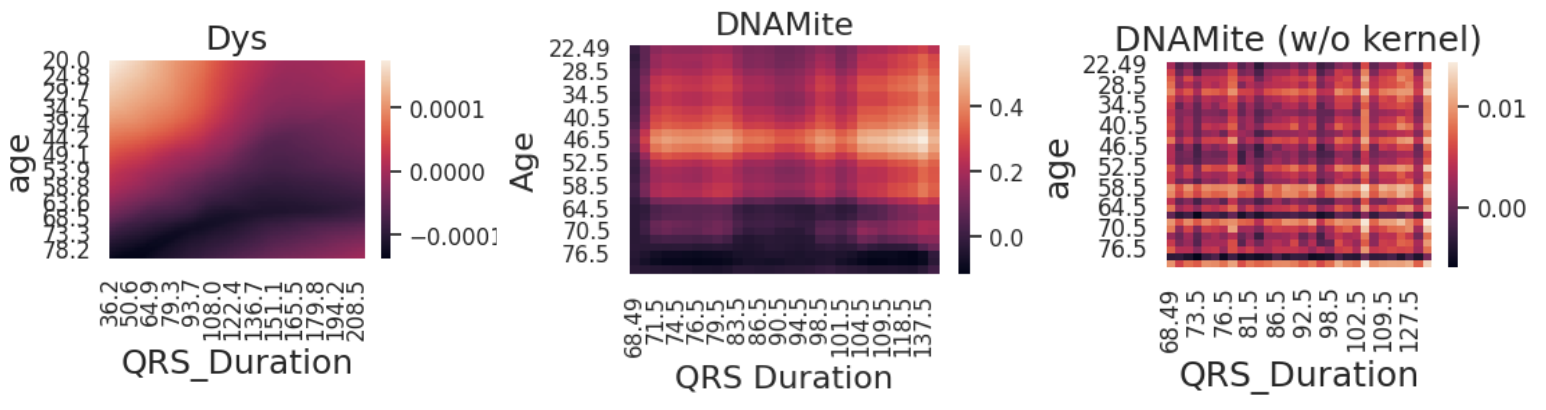}
    \includegraphics[width=\linewidth]{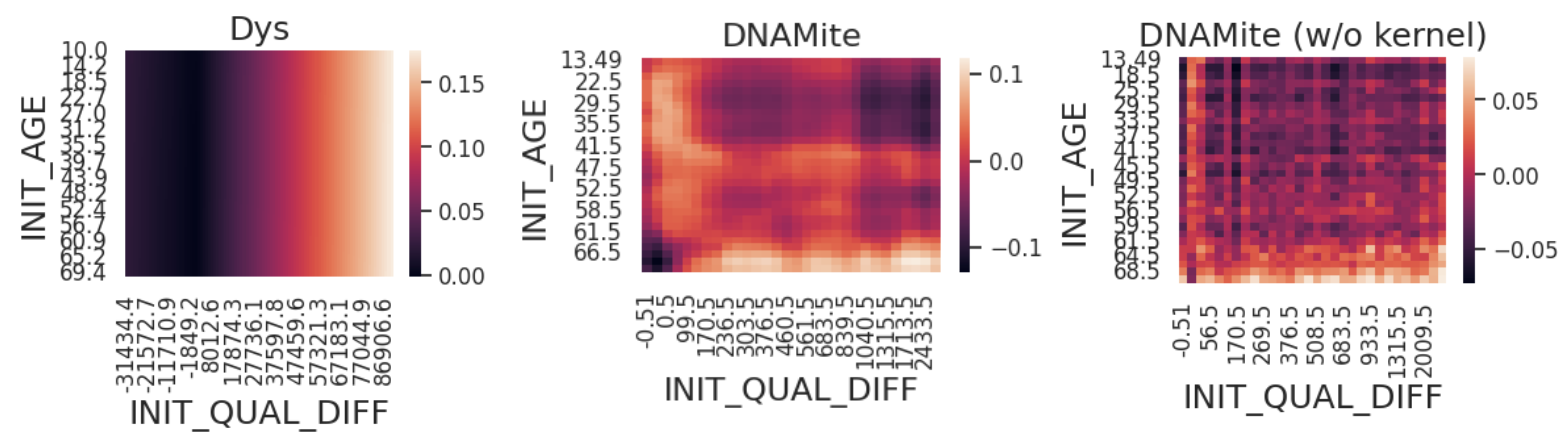}
    \includegraphics[width=\linewidth]{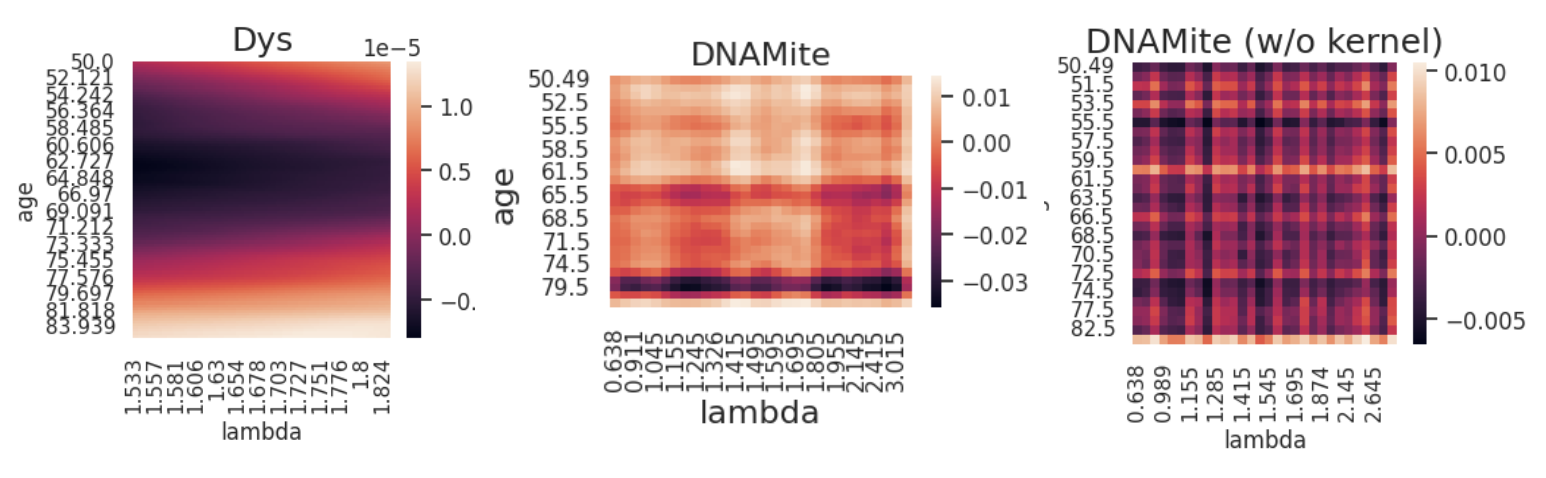}
    \caption{Interaction shape functions to compliment the results in Figure \ref{fig:calibration}. Rows in order: 1) age $||$ QRS duration on heart failure (5 years) 2) INIT\_AGE $||$ INIT\_QUAL\_DIFF on unos (2 years) 3) age $||$ lambda on  flchain (10 years). We don't include plots from support and metabric as we found there were no important interaction features on either dataset.}
    \label{fig:extra_calibration}
\end{figure*}

\end{document}